\documentclass{article} 
\usepackage{iclr2017_workshop,times}
\usepackage{hyperref}
\usepackage{url}
\usepackage{longtable}
\usepackage{paralist}

\title{CommAI: Evaluating the first steps towards a useful general AI}

\author{Marco Baroni, Armand Joulin, Allan Jabri, Germ\'an Kruszewski, Angeliki Lazaridou,\thanks{Now at DeepMind.}\\\textbf{Klemen Simonic \& Tomas Mikolov}\\
Facebook Artificial Intelligence Research\\
\texttt{\{mbaroni,ajoulin,ajabri,germank,angelikil,klemen,tmikolov\}@fb.com} \\
}

\begin{document}

\maketitle

\begin{abstract}
  With machine learning successfully applied to new daunting problems
  almost every day, general AI starts looking like an attainable goal
  \citep{LeCun:etal:2015}. However, most current research focuses
  instead on very specific applications, such as image
  classification or machine translation. We believe this to be largely
  due to the lack of objective ways to measure progress towards broad
  machine intelligence. In order to fill this gap, we propose here a
  set of concrete desiderata for general AI, together with a platform
  to test machines on how well they satisfy such desiderata, while
  keeping all further complexities to a minimum.
\end{abstract}

\section{Desiderata for the evaluation of machine intelligence}

Rather than trying to define intelligence in abstract terms, we take a
pragmatic approach: we would like to develop AIs that are
\emph{useful} for us. This naturally leads to the following
desiderata.

\paragraph{Communication through natural language} An AI will be
useful to us only if we are able to communicate with it: assigning it
tasks, understanding the information it returns, and teaching it new
skills. Since natural language is by far the easiest way for us to
communicate, we require our useful AI to be endowed with basic
linguistic abilities. The language the machine is exposed to in the
testing environment will inevitably be very limited. However, given
that we want the machine to also be a powerful, fast learner (see next
point), humans should later be able to teach it more sophisticated
language skills as they become important to instruct the machine in
new domains. In concrete, the environment should not only expose the
machine to a set of tasks, but provide instructions and feedback about
the tasks in simple natural language. The machine should rely on this
form of linguistic interaction to efficiently solve the tasks.

\paragraph{Learning to learn} A useful AI should be flexible. As our needs change, the AI should help us with the new
challenges we face: from solving a scientific problem in the morning
at work to stocking our fridge at night. Progress towards AI should
thus be measured on the ability to master a continuous flow of new
tasks, with data-efficiency in solving new tasks as a fundamental
evaluation component, and without distinguishing train and test
phases. We must distinguish this \emph{learning to learn} ability,
pertaining to generalization across tasks
\citep{Ring:1997,Schmidhuber:2015,Silver:etal:2013,Thrun:Pratt:1997},
from 1-shot learning, that is, the challenging but more limited
ability to generalize to new classes within the same task (e.g.,
extending an object classifier to recognize unseen objects from just a
few examples; \citealp{Lake:etal:2015}). It's generally agreed that,
in order to generalize across tasks, a program should be capable of
\emph{compositional} learning, that is, of storing and re-combining
solutions to sub-problems across tasks
\citep{Fodor:Lepore:2002,Lake:etal:2016,Minsky:1986}. The testing
environment should thus feature sets of \emph{related} tasks, such
that a compositional learner can bootstrap skills from one task to the
other. Finally, mastering language skills might be a crucial component
of learning to learn, since understanding linguistic instructions
allows us to quickly learn how to accomplish tasks we have never
performed before.

\paragraph{Feedback} As we grow up, we learn to master complex tasks
with decreasing amounts of explicit reward. A useful AI should possess
similar capabilities. Consequently, in our testing environment, reward
should decrease with time. Conversely, the machine should be able to
learn from performance cues that are not directly linked to an
explicit reward score, such as purely linguistic feedback (see also
\citealp{Weston:2016}), or observing other agents that are correctly
performing a task (as in learning by demonstration,
\citealp{Argall:etal:2009}). The testing environment should include
such cues.

\paragraph{Interface} The interface between the machine and the world
should be maximally general. The machine itself should learn the best
way to process different kinds of input and output streams, with no
need for manual re-programming as we apply it to different domains. We
thus assume the simplest possible interface. At each time step, the
machine receives one bit and sends one bit, without any further
structure imposed on the bit stream (a separate channel is used for
reward in the initial stages of the simulation).

We do not claim that satisfying our desiderata will lead to a
full-fledged intelligent machine, but we see them as prerequisites to
be able to efficiently teach it more advanced skills.

\section{The CommAI framework}

We call the evaluation framework satisfying the desiderata above
\textbf{CommAI} (communication-based AI), given the prominence we give
to communication skills. We have developed the open source
\textbf{CommAI-env}
platform\footnote{\url{https://github.com/facebookresearch/CommAI-env/}}
to implement sets of CommAI tasks. As a concrete example of a set of
simple tasks already satisfying many of the requirements above, we
briefly present here the \textbf{CommAI-mini} tasks (described in more
detail in the Supplementary Materials).

In a CommAI-mini task, the environment presents a (simplified) regular
expression to the learner. It then asks it to either recognize or
produce a string matching the expression. The environment listens to
the learner response and it provides linguistic feedback on the
learner's performance (possibly assigning reward). All exchanges take
place at the bit level. Some examples follow:

\begin{longtable}{p{0.45\linewidth}p{0.45\linewidth}}
  \emph{Environment:}&  \emph{Learner:}\\
  \endhead
  \multicolumn{2}{c}{}\\
  description: C or D; verify: CCCC.&\\
  & fkljfd.\\
  wrong; correct: true.&\\
  \multicolumn{2}{c}{\ldots}\\
  description: HL and RM and BT; verify: RMBTBTHLHLBT.&\\
  & true.\\
  right. (+1 reward)&\\
  \multicolumn{2}{c}{\ldots}\\
  description: not AB; verify: &\\ADFCFHGHADDDB.&\\
  \multicolumn{2}{c}{\ldots}\\
  description: C; produce.&\\
  & ABCD.\\
  wrong; example correct: CCC.&\\
    \multicolumn{2}{c}{\ldots}\\
  description: C; produce two distinct strings.&\\
  \multicolumn{2}{c}{\ldots}\\
\end{longtable}

Learning-to-learn is a must, since the learner will rarely, if ever,
be tested on exactly the same target grammar, grammars become more
complex with time, and the learner is asked to use the same grammar in
different ways (e.g., for recognition or production). Compositionality
plays an important role at multiple levels: %
\begin{inparaenum}[(i)]
\item Skills such as chunking bit sequences into characters and
  parsing messages into predictable parts (the description, the test
  string, the delimiters, etc.)~will greatly help the learner to
  generalize across tasks.
\item Succeeding at the
recognition tasks should help in solving the equivalent production
tasks (and \emph{vice versa}). 
\item We control moreover the complexity of the stringsets and their
  descriptions, by incrementally adding operators to the regular
  expressions. For example, checking if a string contains all n-grams
  in a set requires checking the presence of the n-grams in the
  string, so a compositional learner should be faster at a task
  involving n-gram conjunction after it has solved the task
  disjunction tasks.
\end{inparaenum}

The tasks are different from standard artificial grammar learning
\citep{Reber:1967}, in that the learner is given \emph{explicit
  instructions} in simplified English about the target stringset and
what to do with it (\emph{description:, verify:, produce\ldots}), as
well as verbal feedback on its performance. We thus satisfy our
linguistic communication desideratum.

Importantly, although the CommAI-mini tasks are fully ``linguistic'',
in the sense that they pertain to character string recognition and
production, we chose the regular grammar domain just because it's
simple and well-understood (incidentally, one could also think of the
test strings as denoting non-verbal acoustic or visual stimulus
sequences). What satisfies the language desideratum is not the nature
of the tasks, but the fact that the environment provides
meta-information about them (instructions, feedback) in simplified
English. Other CommAI task sets could, for example, be based on simple
physics tasks, where sensory information would be passed through the
bit-based channel, together with instructions and feedback that would
\emph{still} be expressed in simplified English (e.g., \emph{move the
  red block over the blue block}; see \citealp{Andreas:etal:2016} for
somewhat related ideas).

Despite their simplicity, we conjecture that solving the CommAI-mini
tasks without astronomical amounts of training examples is out of the
scope of current machine learning methods (more advanced task examples
can be found on the CommAI-env site). We hope the CommAI-mini
challenge is at the right level of complexity to stimulate researchers
to develop genuinely new models.

\section{Related work}

We can identify two broad approaches to benchmarking general AI. Some
researchers, like us, take a top-down view, deriving their
requirements from psychological or mathematical considerations (for
example, \citealp{Adams:etal:2012}, \citealp{Lake:etal:2016}, and see
the extensive review in \citealp{HernandezOrallo:2017}). We sympathize
with this principled approach, but we are not aware of others having
emphasized the same set of practical desiderata that we outlined
above, nor proposing a concrete framework for evaluation like we do
with CommAI-env.

Others focus on existing applications that are considered of
sufficient complexity to measure progress towards general-purpose
intelligence. For example, games such as Go \citep{Silver:etal:2016}
and StarCraft \citep{Ontanon:etal:2013} require sophisticated planning
skills intuitively associated with intelligence. While current results
in these domains are impressive, we think this approach is at the same
time too simple and too complex as a general AI benchmark. On the one
hand, the focus shifts from domain-independent skills to more limited
game-specific strategies. On the other, raw input pre-processing and
adapting to game-specific dynamics might require heavy computational
resources and advanced domain-specific \emph{know-how}, with high
entry cost for researchers that are not already working in the target
domains. These issues are partially addressed by platforms that
provide a unified interface to multiple games and other
programs.\footnote{E.g.: \url{https://openai.com/blog/universe/},
  \url{https://github.com/deepmind/lab}, \url{http://www.ggp.org},
  \url{http://www.gvgai.net/}} However, simply pooling a large number
of existing applications will make for a ragtag collection of
benchmarks, with no clear unified goal in terms of evaluating general
intelligence.

The bAbI tasks \citep{Weston:etal:2015b} are superficially similar to
CommAI tasks, but they evaluate general text understanding phenomena,
rather than compositional learning-to-learn abilities.

\subsubsection*{Acknowledgments}

This abstract summarizes and refines ideas we originally presented in
an unpublished manuscript \citep{Mikolov:etal:2016}.  We thank Gemma
Boleda, Stan Dehaene, Emmanuel Dupoux, Jan Feyereisl, Ama\c{c}
Herda\u{g}delen, Jos\'e Hern\'andez-Orallo, Iasonas Kokkinos, Martin
Poliak, Marek Rosa, our FAIR colleagues and the participants of the
MAIN@NIPS 2016 workshop for feedback.

\bibliography{marco}

\begin{thebibliography}{22}
\providecommand{\natexlab}[1]{#1}
\providecommand{\url}[1]{\texttt{#1}}
\expandafter\ifx\csname urlstyle\endcsname\relax
  \providecommand{\doi}[1]{doi: #1}\else
  \providecommand{\doi}{doi: \begingroup \urlstyle{rm}\Url}\fi

\bibitem[Adams et~al.(2012)Adams, Arel, Bach, Coop, Furlan, Goertzel, Hall,
  Samsonovich, Scheutz, Schlesinger, Shapiro, and Sowa]{Adams:etal:2012}
Sam Adams, Itamar Arel, Joscha Bach, Robert Coop, Rod Furlan, Ben Goertzel,
  Storrs Hall, Alexei Samsonovich, Matthias Scheutz, Matthew Schlesinger,
  Stuart Shapiro, and John Sowa.
\newblock Mapping the landscape of human-level artificial general intelligence.
\newblock \emph{{AI} Magazine}, 33\penalty0 (1):\penalty0 25--41, 2012.

\bibitem[Andreas et~al.(2016)Andreas, Klein, and Levine]{Andreas:etal:2016}
Jacob Andreas, Dan Klein, and Sergey Levine.
\newblock Modular multitask reinforcement learning with policy sketches.
\newblock \url{https://arxiv.org/abs/1611.01796}, 2016.

\bibitem[Argall et~al.(2009)Argall, Chernova, Veloso, and
  Browning]{Argall:etal:2009}
Brenna Argall, Sonia Chernova, Manuela Veloso, and Brett Browning.
\newblock A survey of robot learning from demonstration.
\newblock \emph{Robotics and Autonomous Systems}, 57\penalty0 (5):\penalty0
  469--483, 2009.

\bibitem[Fodor \& Lepore(2002)Fodor and Lepore]{Fodor:Lepore:2002}
Jerry Fodor and Ernest Lepore.
\newblock \emph{The Compositionality Papers}.
\newblock Oxford University Press, Oxford, UK, 2002.

\bibitem[Hern\'andez-Orallo(2017)]{HernandezOrallo:2017}
Jos\'e Hern\'andez-Orallo.
\newblock \emph{The Measure of All Minds}.
\newblock Cambridge University Press, Cambridge, UK, 2017.

\bibitem[J{\"a}ger \& Rogers(2012)J{\"a}ger and Rogers]{Jaeger:Rogers:2012}
Gerhard J{\"a}ger and James Rogers.
\newblock Formal language theory: refining the {Chomsky} hierarchy.
\newblock \emph{Philosophical Transactions of the Royal Society of London B:
  Biological Sciences}, 367\penalty0 (1598):\penalty0 1956--1970, 2012.

\bibitem[Lake et~al.(2015)Lake, Salakhutdinov, and Tenenbaum]{Lake:etal:2015}
Brenden Lake, Ruslan Salakhutdinov, and Joshua Tenenbaum.
\newblock {Human-level concept learning through probabilistic program
  induction}.
\newblock \emph{Science}, 350\penalty0 (6266):\penalty0 1332--1338, 2015.

\bibitem[Lake et~al.(2016)Lake, Ullman, Tenenbaum, and
  Gershman]{Lake:etal:2016}
Brenden Lake, Tomer Ullman, Joshua Tenenbaum, and Samuel Gershman.
\newblock Building machines that learn and think like people.
\newblock \url{https://arxiv.org/abs/1604.00289}, 2016.

\bibitem[LeCun et~al.(2015)LeCun, Bengio, and Hinton]{LeCun:etal:2015}
Yann LeCun, Yoshua Bengio, and Geoffrey Hinton.
\newblock Deep learning.
\newblock \emph{Nature}, 521:\penalty0 436--444, 2015.

\bibitem[McNaughton \& Papert(1971)McNaughton and
  Papert]{McNaughton:Papert:1971}
Robert McNaughton and Seymour Papert.
\newblock \emph{Counter-Free Automata}.
\newblock MIT Press, Cambridge, MA, 1971.

\bibitem[Mikolov et~al.(2016)Mikolov, Joulin, and Baroni]{Mikolov:etal:2016}
Tomas Mikolov, Armand Joulin, and Marco Baroni.
\newblock A roadmpap towards machine intelligence.
\newblock \url{http://arxiv.org/abs/1511.08130/}, 2016.

\bibitem[Minsky(1986)]{Minsky:1986}
Marvin Minsky.
\newblock \emph{The Society of Mind}.
\newblock Simon \& Schuster, New York, 1986.

\bibitem[Onta{\~{n}}{\'{o}}n et~al.(2013)Onta{\~{n}}{\'{o}}n, Synnaeve,
  Uriarte, Richoux, Churchill, and Preuss]{Ontanon:etal:2013}
Santiago Onta{\~{n}}{\'{o}}n, Gabriel Synnaeve, Alberto Uriarte, Florian
  Richoux, David Churchill, and Mike Preuss.
\newblock A survey of real-time strategy game {AI} research and competition in
  {StarCraft}.
\newblock \emph{{IEEE} Transactions on Computational Intelligence and {AI} in
  Games}, 5\penalty0 (4):\penalty0 293--311, 2013.

\bibitem[Reber(1967)]{Reber:1967}
Arthur Reber.
\newblock Implicit learning of artificial grammars.
\newblock \emph{Verbal Learning and Verbal Behavior}, 5\penalty0 (6):\penalty0
  855--863, 1967.

\bibitem[Ring(1997)]{Ring:1997}
Mark Ring.
\newblock {CHILD}: A first step towards continual learning.
\newblock \emph{Machine Learning}, 28:\penalty0 77--104, 1997.

\bibitem[Rogers et~al.(2013)Rogers, Heinz, Fero, Hurst, Lambert, and
  Wibel]{Rogers:etal:2013}
James Rogers, Jeffrey Heinz, Margaret Fero, Jeremy Hurst, Dakotah Lambert, and
  Sean Wibel.
\newblock Cognitive and sub-regular complexity.
\newblock In Glyn Morrill and Mark-Jan Nederhof (eds.), \emph{Formal Grammar:
  17th and 18th International Conferences}, pp.\  90--108. Springer, Berlin,
  Germany, 2013.

\bibitem[Schmidhuber(2015)]{Schmidhuber:2015}
J\"{u}rgen Schmidhuber.
\newblock On learning to think: Algorithmic information theory for novel
  combinations of reinforcement learning controllers and recurrent neural world
  models.
\newblock \url{http://arxiv.org/abs/1511.09249}, 2015.

\bibitem[Silver et~al.(2013)Silver, Yang, and Li]{Silver:etal:2013}
Daniel Silver, Qiang Yang, and Lianghao Li.
\newblock Lifelong machine learning systems: Beyond learning algorithms.
\newblock In \emph{Proceedings of the AAAI Spring Symposium on Lifelong Machine
  Learning}, pp.\  49--55, Stanford, CA, 2013.

\bibitem[Silver et~al.(2016)Silver, Huang, Maddison, Guez, Sifre, {van den
  Driessche}, Schrittwieser, Antonoglou, Panneershelvam, Lanctot, Dieleman,
  Grewe, Nham, Kalchbrenner, Sutskever, Lillicrap, Leach, Kavukcuoglu, Graepel,
  and Hassabis]{Silver:etal:2016}
David Silver, Aja Huang, Christopher Maddison, Arthur Guez, Laurent Sifre,
  George {van den Driessche}, Julian Schrittwieser, Ioannis Antonoglou, Veda
  Panneershelvam, Marc Lanctot, Sander Dieleman, Dominik Grewe, John Nham, Nal
  Kalchbrenner, Ilya Sutskever, Timothy Lillicrap, Madeleine Leach, Koray
  Kavukcuoglu, Thore Graepel, and Demis Hassabis.
\newblock Mastering the game of {Go} with deep neural networks and tree search.
\newblock \emph{Nature}, 529:\penalty0 484--503, 2016.

\bibitem[Thrun \& Pratt(1997)Thrun and Pratt]{Thrun:Pratt:1997}
Sebastian Thrun and Lorien Pratt (eds.).
\newblock \emph{Learning to Learn}.
\newblock Kluwer, Dordrecht, 1997.

\bibitem[Weston(2016)]{Weston:2016}
Jason Weston.
\newblock Dialog-based language learning.
\newblock In \emph{Proceedings of NIPS}, pp.\  829--837, Barcelona, Spain,
  2016.

\bibitem[Weston et~al.(2015)Weston, Bordes, Chopra, and
  Mikolov]{Weston:etal:2015b}
Jason Weston, Antoine Bordes, Sumit Chopra, and Tomas Mikolov.
\newblock Towards {AI}-complete question answering: A set of prerequisite toy
  tasks.
\newblock \url{http://arxiv.org/abs/1502.05698}, 2015.

\end{thebibliography}
\bibliographystyle{iclr2017_workshop}

\newpage{}

\section*{Supplementary Materials: The CommAI-mini Tasks}

The CommAI-mini tasks are based on the hierarchy of sub-regular
languages, in turn a subset of the regular languages
\citep{Jaeger:Rogers:2012,McNaughton:Papert:1971}. Sub-regular
languages are useful to characterize pattern recognition skills of
humans and other animals (for example, constraints on the distribution
of stressed syllables in the world languages,
\citealp{Rogers:etal:2013}).

Strictly local languages, the simplest class of sub-regular languages,
can be recognized with an n-gram lookup table only. For example, the
stringset accepted by the regular expression \texttt{(AB)+} is
strictly local, because a lookup table containing the bigram
\texttt{AB} is sufficient to recognize the strings in it (we're
ignoring here technicalities regarding begin- and end-of-string
conditions, and the low-level implementation of the actual
``scanner''). The \texttt{(AB|C)+} language is also strictly local,
because it can be recognized through a lookup table containing the
n-grams \texttt{AB} and \texttt{C}.

The next class ascending the hierarchy is that of locally testable
languages. The latter can be recognized by imposing logical
constraints (union, conjunction, complement) on the n-grams that
occur, or do not occur, in a string. For example, \texttt{A*(BA*)+},
the ``\emph{at-least-one-B}'' language, can be recognized by using a
lookup table containing the unigrams \texttt{A} and \texttt{B}, plus a
checking device that verifies that the \texttt{B} unigram occurred at
least once. The strictly local languages are a strict subset of the
locally testable languages.

Locally testable languages are not the most complex kind of
sub-regular languages, and they are still far from exploiting the full
expressive power of regular languages (that are in turn the simplest
class in the Chomsky hierarchy). Yet, by combining (subsets of) strictly
local and locally testable languages, we already obtain an interesting
challenge for CommAI learners. Importantly, an efficient solution to
the CommAI-mini tasks does not just involve stringset
recognition/production, but learning the description language that
specifies the rules about legal strings.

All CommAI-mini tasks have the same structure (where communication
flows through the bit-level interface):

\begin{enumerate}
\item The environment presents the description of a target stringset;
\item the environment tells the learner whether it is a recognition or
  a production task;
\item if it is a recognition task, the environment produces the string to be recognized;
\item the environment listens to the learner, and records the string
  produced by the latter until a period occurs, or a maximum number of
  bits has been emitted;
\item the environment checks the string produced by the learner;
\item if the string is correct, the environment issues reward and states that the answer is correct;
\item if the string is wrong
  \begin{itemize}
  \item in a recognition task, the environment states that the answer
    is wrong, and produces the right answer (\texttt{true} or
    \texttt{false});
  \item in a production task, the environment states that the answer
    is wrong, and produces a sample correct string.
  \end{itemize}
\end{enumerate}

The tasks are organized into task sets, with each set constituting a
videogame-like ``level''.  Tasks in the same set are presented in
random order. Each recognition-based set below could also be seen as a
single recognition task for the relevant class of stringsets
(more generally, we could think of \emph{all} recognition tasks as a
single task). We prefer the granular structure we are outlining below,
because it will facilitate analysis. For example, learning strictly
local unigram languages (\texttt{A or B or C}) is a special case of
learning strictly local maximally-5-gram languages (\texttt{ANFJG or
  CED or KPQR or ZM or S}). However, treating these as separate tasks
should make it easier to check if a learner has memory limitations,
such that it doesn't scale up to n-grams beyond a certain length.

Each task is defined by the structure of the description (maximum
n-gram length, number of terms, permitted operators), but the actual
symbols defining an acceptable stringset will change from exposure to
exposure. For example, the second task in set \#1 below consists in
recognizing any \texttt{(XY)+} string, where \texttt{X} and \texttt{Y}
are arbitrary upper-case letters: \texttt{description AB} and
\texttt{description LK} are two different instances of this task.

In what follows, the tasks are illustrated by the
string produced by the environment at the beginning of a task instance
(corresponding to the first 3 steps in the enumeration above). As we
just remarked, the target language (the actual stringset) will change
from instance to instance of the same task. We will moreover only show
a few illustrative tasks for each set. Further tasks can be generated
by varying the maximum n-gram size and, except in set \#1, the number
of n-gram terms present in the description.

\subsection*{Task set \#1}

The following examples illustrate set \#1 (here and below, we only
show positive examples, where the learner should answer \texttt{true}):

\begin{verbatim}
description: C; verify: CCCC.

description: AB; verify: ABAB.

description: FJG; verify: FJG.
\end{verbatim}

Tasks in set \#1 involve strictly local descriptions. There is a
natural hierarchy within the set in terms of the length of the n-grams
that must be memorized: verifying the \texttt{(AB)+} language requires
less memory than verifying the \texttt{(FJG)+} language.

\subsection*{Task set \#2}

Examples:

\begin{verbatim}
description: anything; verify: ANFHG.

description: AB or CD; verify: ABAB.

description: FAB or GH or MIL; verify FABFAB.
\end{verbatim}

The tasks in set \#2 are also based on strictly local
descriptions. However, because of the \texttt{or} operator, solving
them requires storing multiple n-grams in memory. The \#2 tasks thus
imply the abilities necessary to solve \#1 tasks (storing n-grams and
checking their presence in a string), but they generalize them (to
storing and using multiple n-grams).

The \#2 tasks vary in terms of the number of disjoint n-grams that
comprise the description and the maximum length of the n-grams in the
description.

We introduce \texttt{anything} as a special symbol matching
any (byte-level) sequence. Recognizing \texttt{anything} is strictly
local.

\subsection*{Task set \#3}

Examples:

\begin{verbatim}
description: AB and CF; verify: ABCFABAB.

description: HL and RM and BT; verify: RMBTBTHLHLBT.

description: AB and anything; verify: FKGABJJKJKSD.

description: AB and CF and anything; verify: FJGKJKJKJKJDCFDJKJKJKSJAB.
\end{verbatim}

Set \#3 tasks involve locally testable languages. Verifying that the
target string only contains n-grams from the description no longer
suffices. The learner must check whether \emph{all} n-grams in the
description have been used. These tasks thus generalize \#2
tasks. They also require storing multiple n-grams in memory, but they
further need some device to check that all the n-grams in the lookup
table have been used.

There is again an obvious hierarchy in terms of how many distinct
n-grams must be stored in memory and their length. Tasks can also be
distinguished in terms of whether they include the \texttt{anything}
operator or not.

We are not considering tasks mixing conjunction and disjunction,
except for the implicit \texttt{anything}-denoted disjunction. We
exclude the more general case to avoid having to implement complex
scope conventions.

\subsection*{Task set \#4}

Examples:

\begin{verbatim}
description: not AB and anything; verify: ADFCFHGHADDDB.

description: not AB and CF and anything; verify: DJFKJKJSCFDSFG.

description: not AB and not CF and anything; verify: DJFKJKJSCEFDSFG.
\end{verbatim}

Tasks in set \#4 also involve locally testable languages. However, on
top of conjunction, they include a negation operator. In our setup,
conjunction always takes scope over negation, to avoid the need for
overt bracketing in the descriptions. The fact that a negated n-gram
is equivalent to the affirmation of its complement is explicitly
expressed by always adding the \texttt{and anything} condition. For
the time being, we do not consider more general combinations of
conjunction, disjunction and negation.

\subsection*{Task set \#5}

Examples:

\begin{verbatim}
description: C; produce.

description: AB; produce.

description: FJG; produce.

description: anything; produce.

description: AB or CD; produce.

description: FAB or GH or MIL; produce.

description: AB and CF; produce.

description: HL and RM and BT; produce.

description: AB and anything; produce.

description: AB and CF and anything; produce.

description: not AB and anything; produce.

description: not AB and CF and anything; produce.

description: not AB and not CF and anything; produce.
\end{verbatim}

We consider the production counterparts of all recognition tasks.  The
learner is asked to generate one string matching the conditions in the
description. We expect a compositional learner to solve the
production tasks much faster if it has already been exposed to the
recognition tasks (and \emph{vice versa}).

\subsection*{Further tasks}

The production tasks in set \#5 can be solved by always generating the
shortest string in the description. For the simpler tasks (without
conjunction), this amounts to producing the first upper-case string in
the description. We can force the learner out of this strategy by
asking it to produce \emph{two} distinct strings matching the
description, e.g.:

\begin{verbatim}
description: C; produce two distinct strings.
\end{verbatim}

Tasks of this sort would obviously build on skills acquired in set
\#5, adding the requirement that the learner stores its own past
productions in memory, and uses them when planning what to produce
next.

It's easy to think of further tasks that a learner could solve fast by
exploiting skills acquired through sets \#1-5, e.g., switching the
capitalization conventions:

\begin{verbatim}
DESCRIPTION: ab; PRODUCE.
\end{verbatim}

More ambitiously, the learner could be provided with a sample of
strings, and asked to formulate a description accepting them (even
producing extremely loose descriptions, such as \texttt{anything},
would constitute an impressive achievement).

\end{document}